\documentclass{article}
\usepackage{spconf,amsmath,graphicx}
\usepackage{amssymb}
\usepackage{booktabs}       
\usepackage{nicefrac}       
\usepackage{url}            
\usepackage{hyperref}
\usepackage{tikz}
\usepackage{tikzit}
\usepackage{pgfplots}
\usepackage{array,multirow}
\tikzset{%
    state/.style = {%
        draw, circle, minimum size = 4, inner sep = 0, fill = black
    }%
}%

\tikzset{%
    dashedarrow/.style = {%
        draw, densely dashed, > = {Latex[width = 1.7mm, length = 2.2mm, open, fill = white]},
        decoration={markings, mark=at position 1 with {\begin{pgfonlayer}{arrowlayer}\arrow{>}\end{pgfonlayer}}},
        postaction={decorate}
    }%
}%

\tikzset{%
  dots/.style args={#1per #2}{%
    line cap=round,
    dash pattern=on 0.5 off 0.5,
  }
}

\DeclareMathOperator*{\argmin}{arg\,min}

\title{Disambiguation of One-Shot Visual Classification Tasks:\\A Simplex-Based Approach}%
\name{Yassir BENDOU, Lucas DRUMETZ, Vincent GRIPON, Giulia LIOI and Bastien PASDELOUP}
\address{IMT Atlantique, Lab-STICC, UMR CNRS 6285, Brest F-29238, France}
\begin{document}
\maketitle
\begin{abstract}
The field of visual few-shot classification aims at transferring the state-of-the-art performance of deep learning visual systems onto tasks where only a very limited number of training samples are available.
The main solution consists in training a feature extractor using a large and diverse dataset to be applied to the considered few-shot task.
Thanks to the encoded priors in the feature extractors, classification tasks with as little as one example (or ``shot'') for each class can be solved with high accuracy, even when the shots display individual features not representative of their classes.
Yet, the problem becomes more complicated when some of the given shots display multiple objects.
In this paper, we present a strategy which aims at detecting the presence of multiple and previously unseen objects in a given shot.
This methodology is based on identifying the corners of a simplex in a high dimensional space. 
We introduce an optimization routine and showcase its ability to successfully detect multiple (previously unseen) objects in raw images.
Then, we introduce a downstream classifier meant to exploit the presence of multiple objects to improve the performance of few-shot classification, in the case of extreme settings where only one shot is given for its class.
Using standard benchmarks of the field, we show the ability of the proposed method to slightly, yet statistically significantly, improve accuracy in these settings.
\end{abstract}
\section{Introduction}
In the field of few-shot learning, the main aim is to reconcile the remarkable performance of deep learning systems, usually obtained thanks to large amount of training data, with the constraint of having a very small number of labeled examples per class (typically 1-5). The main strategy relies on the use of transfer learning, where feature extractors are trained using large and diverse datasets, to be applied as a first step onto the considered few-shot task. In the case of vision systems and classification tasks, high accuracy can be achieved even when using very simple classifiers on top of the extracted feature vectors~\cite{wang2019simpleshot}. In this work, we focus on the extreme case where a single training sample is available for each considered class.

\begin{figure}[]
\begin{center}
    \includegraphics[scale=0.35]{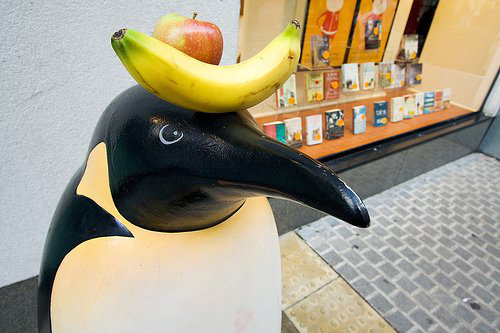}
    \caption{Image extracted from the MiniImageNet benchmark for visual few-shot classification. This image was automatically found to be the one in the benchmark that has the most negative impact to performance when it appears in randomly sampled few-shot tasks. Indeed, even though it contains multiple objects in the foreground, the class it is associated with is ``library'', creating dramatic ambiguities when it is selected as the only example from that class.}
    \label{fig:penguin}
\end{center}
\end{figure}

A straight limitation to this strategy comes from the fact that natural images usually contain multiple objects, as depicted in Figure~\ref{fig:penguin}. As a matter of fact, using a simple global feature vector to summarize such images is likely to average features of all depicted objects, or even worse, disregard some of them. A direct consequence is a likely significant drop in accuracy compared to what could be achievable if such objects could be treated independently from one another.

The main purpose of this contribution is to propose an automatic methodology to detect multiple (previously unseen) objects in an image, using the same premises as the usual few-shot classification framework. To do so, we introduce a statistical modeling of feature vectors obtained from random crops in an image from any pre-trained feature extractor. We propose an optimization routine to automatically find objects based on this modeling. Using standardized benchmarks of the field of visual few-shot classification, we show the ability of the proposed pipeline to outperform existing methods.

\begin{figure*}[]
\begin{center}
    \scalebox{0.92}{\input{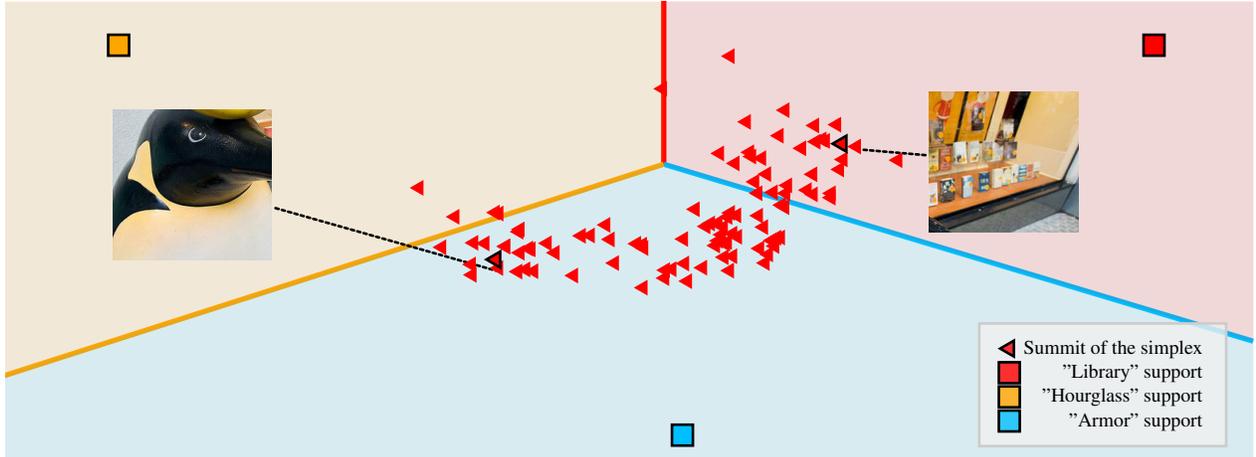}}
\vspace{-4mm}
    \caption{Representation of the feature vectors associated with 200 random crops from the image depicted in Figure~\ref{fig:penguin}. The feature extractor used here, a ResNet-12 from~\cite{resnet}, outputs vectors with 640 dimensions. For visualization purposes, we chose to project these high-dimensional vectors onto a 2d-plane containing the centroids for the classes "armor", "hourglass" and "library", the three classes in the MiniImageNet dataset that are the closest from these feature vectors. We also depicted in color the Voronoi cells corresponding to these three classes. We see that the feature vectors display an interesting distribution following what resembles an arc, starting from crops close to the "library" class on the top right to crops close to the "armor" class on the bottom left. The vertices circled in black correspond to those automatically found using the methodology described in~\ref{subsection:simplex_extraction}. The images corresponding to the closest crops to these vertices are displayed. We observe that both the object ``penguin'' and the object ``library" have been correctly retrieved from this example.}
    \label{fig:QR}
\end{center}
\end{figure*}

The outline of the paper is as follows. In Section~\ref{rw}, we introduce related work. In Section~\ref{methodo}, we introduce our mathematical modeling and the optimization problem to automatically find objects in a given input image. In Section~\ref{results} we display results on classical benchmarks in the field and analyze the importance of introduced hyperparameters. Section~\ref{conclusion} is a conclusion.
\section{Related work}
\label{rw}

Few-shot classification has become a very popular problem in the past few years. We can group contributions into two main lines of work:
\begin{itemize}
    \item Optimization-based approaches such as meta-learning which aim to effectively adapting a model parameters to new few-shot tasks. For instance, MAML and many of its variants~\cite{Finn2017, MetaOpt, liu2020task} aim to find a good model initialization to adapt to new tasks with few gradient updates;
    \item Transfer-based approaches, where a model is pre-trained to find an appropriate feature space where each image is well represented for transfer tasks~\cite{wang2019simpleshot, vinyals2016matching, Snell2017, Chen2019}. This category has known a large success in the recent years due to its good performance while remaining relatively simple. Our methodology falls in this category.
\end{itemize}

Training the feature extractor often relies on data augmentation on the base dataset through random rotations or random crops~\cite{mangla2020charting}. Data augmentation on the few-shot dataset, although less frequent, has also been explored. For example, one approach computes the optimal transport between different crops of the support set and the query set in order to better estimate the distance between images~\cite{zhang2020deepemd}. Another approach uses random crops to distinguish between crops in the foreground and the background~\cite{luo2021rectifying}. A different approach simply averages multiple features obtained through randomly cropping the images in order to reduce the influence of undesired objects~\cite{Bendou2022}. Here, we use the methodology described in~\cite{Bendou2022} to train our feature extractors, as it has been shown to reach state-of-the-art accuracy on multiple benchmarks in the field of vision.

\section{Methodology}
\label{methodo}

Let us recall the typical pipeline to solve a visual few-shot classification problem:

\begin{enumerate}
\item The first step is to train a feature extractor $f$, usually a deep learning architecture, using a large generic dataset.
\item Then, the images $\mathbf{x}$ from the considered few-shot tasks are transformed into associated feature vectors $f(\mathbf{x})$.
\item Finally, a classifier is trained using the couples $(f(\mathbf{x}), y)$ from the training set, $y$ is the class associated with $\mathbf{x}$.
\end{enumerate}

Our proposed method consists in changing steps 2. and 3. from the previous description. Namely, we aim at automatically detecting the presence of multiple objects when computing the feature vectors in step 2. (Section~\ref{subsection:simplex_extraction}), and adapting the classifier in step 3 (Section~\ref{subsection:classification}), so that it tries to match an object from the training samples with an object from the query image.

Since we are working with images, we suppose that each object in the input image is associated with a specific crop. To better illustrate this, let us consider the image of Figure~\ref{fig:penguin}. By generating random crops in the image, we obtain various feature vectors. The distribution of these feature vectors is complex, as most feature extractors will output high dimensional vectors.

In Figure~\ref{fig:QR}, we show the projection of these feature vectors when using the ResNet-12 feature extractor trained in~\cite{Bendou2022} onto a 2d-plane. This 2d-plane is the one that contains the centroids of the three closest classes in the corresponding MiniImageNet dataset: ``armor", ``hourglass" and ``library".

Because some crops may contain a single object and other multiple ones, we assume the distribution in general to take the form of a simplex whose vertices would be the feature vectors of the objects in the scene. Under this hypothesis, any feature vector associated with a crop can be expressed as a linear interpolation based on the importance of each object in the selected crop.

\subsection{Vertices extraction}
\label{subsection:simplex_extraction}
Let us introduce some notations. We model the distribution of each image in the feature space as a simplex with $K$ vertices, with $K$ assumed fixed for now. We will see in Section~\ref{Choosing the number of vertices} how to identify the right number of vertices for each image. Let $N$ be the number of crops for each image. Let $D$ be the dimension of a feature vector (typically a few hundreds). We define $\mathbf{X} \in \mathcal{M}_{N \times D}(\mathbb{R})$ the matrix containing all the features vectors of crops of the image. Let $\mathbf{D} \in \mathcal{M}_{K \times D}(\mathbb{R})$ be the matrix containing the vertices of the simplex and $\mathbf{W} \in \mathcal{M}_{N \times K}(\mathbb{R})$ the matrix containing the weighting of each vertex to each feature vector. We note $\mathbf{1}_N$ the vector of dimension $N$ filled with 1.

The first part of our optimization problem consists in reconstructing our points from the $K$ unknown vertices. This amounts to solving the following equation: 
\begin{equation}
\begin{aligned}
\mathbf{D}^{*}, \mathbf{W}^{*} = \argmin_{\substack{
    \mathbf{D}, \mathbf{W} 
    }}\| \mathbf{W} \mathbf{D} - \mathbf{X} \|_2^2\\
    \text{s.t} \quad
        0\leq \mathbf{W} \quad \text{and} \quad 
        \mathbf{W}\mathbf{1}_K=\mathbf{1}_N.
\end{aligned}
\label{eq:MSE}
\end{equation}

This form is very common in source separation ~\cite{Dias2012,Drumetz2020}, and can be solved by alternating between estimating $\mathbf{D}$ and $\mathbf{W}$. Such a problem is not convex in both $\mathbf{D}$ and $\mathbf{W}$ and has an infinite number of solutions (it suffices to take vertices far enough apart to perfectly encompass the data), which makes it ill-posed. Among the potential solutions, we are interested in those whose vertices are relatively close to the data. Indeed, these should correspond approximately to possible crops of the input images. It is therefore necessary to add a regularization term to limit the scattering of the vertices in the feature space. Among the possible regularizations, some methods try to minimize the volume of the simplex. As such an objective is complex to achieve, a good approximation is to use a convex relaxation such as minimizing the sum of the squared distances between all possible pairs of vertices, as follows:

\begin{equation}
\begin{aligned}
\mathbf{D}^{*}, \mathbf{W}^{*} = \argmin_{\substack{
    \mathbf{D}, \mathbf{W} 
    }}
    \left(
    \begin{array}{l}
        \lambda \sum\limits_{k=1}^{K-1}\sum\limits_{k'=k+1}^{K} \|\mathbf{D}_k-\mathbf{D}_{k'} \|_2^2 \\
        + (1-\lambda) \| \mathbf{W} \mathbf{D} - \mathbf{X} \|_2^2
    \end{array}
    \right) 
    \\
    \text{s.t} \quad
        0\leq \mathbf{W} \quad \text{and} \quad 
        \mathbf{W}\mathbf{1}_K=\mathbf{1}_N,
\end{aligned}
\label{eq:Equation Optimization}
\end{equation}
where $\mathbf{D}_k$ corresponds to the $k$-th column of $\mathbf{D}$, and $\lambda \in [0, 1]$ is a regularization hyperparameter fixed at $0.05$. The problem is convex in $\mathbf{D}$ and in $\mathbf{W}$, but not simultaneously in both variables. In order to solve this optimization problem, we proceed by alternating minimizations. We use the method proposed by ~\cite{Berman2004} which uses a closed form to solve for $\mathbf{D}$. In order to enforce the constraints of the problem, the resolution of $\mathbf{W}$ is done through gradient descent on a matrix $\mathbf{V} \in \mathcal{M}_{N \times K}(\mathbb{R})$ such that $\forall k : \mathbf{W}_k = \frac{\exp(\mathbf{V}_k)}{\sum_{k' = 1}^K \exp(\mathbf{V}_{k'})}$. Moreover, the initialization of the vertices of the matrix $\mathbf{D}$ is done by choosing $K$ random vectors of the matrix $\mathbf{X}$. The entries of $\mathbf{V}$ are initialized uniformly at random according to a $\mathcal{N}(0, 1)$ distribution. The solution is a stationary point of the cost function with no guarantee of global optimality.

\subsection{Choosing the number of vertices}
\label{Choosing the number of vertices}
So far, we have considered $K$ known and fixed. However, the number of objects of interest in an image is not known in practice.
In this section, we propose an automatic method to identify this number. In order to simplify the analysis, we exploit the fact that a natural image often contains a small number of objects of interest ($K \leq 3)$. This is confirmed empirically by visualizing the distribution of random crops of different images in the feature space. Next, we exploit the reconstruction error of equation ~\eqref{eq:MSE} to establish a criterion for selecting the number of vertices, similar to the one used in the elbow method for selecting the number of clusters in the $K$-means algorithm. The idea is that if the reconstruction error does not decrease significantly ($1.5$ ratio threshold) when increasing the number of vertices, then the previous number is retained.
This method is applied on the entire dataset to identify the number of objects present in each image.
\begin{figure*}[!t]
    \centering
    \scalebox{0.98}{\input{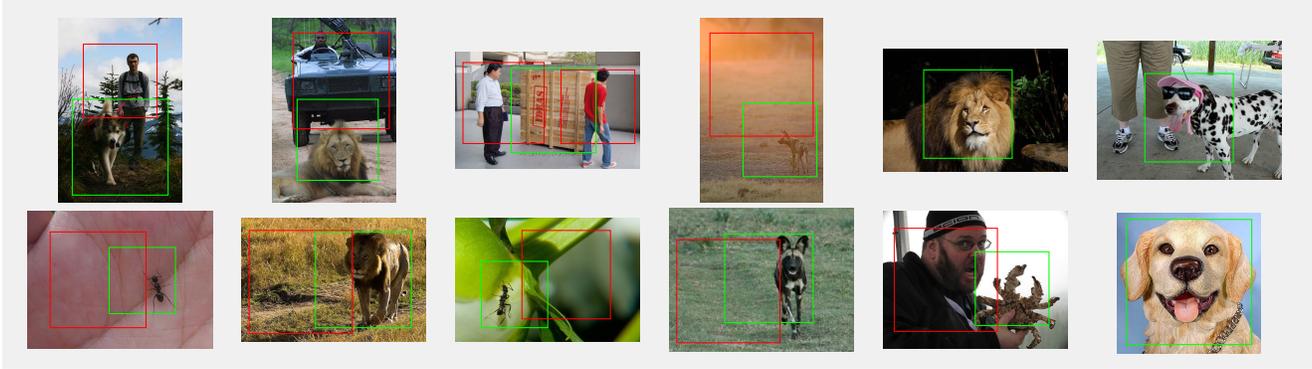}}
    \vspace{-0.3cm}
    \caption{Results of our simplex method on different images from Mini-ImageNet. The frames represent the closest crops to the vertices of the obtained simplexes. We use the green color for the crops that best fit the corresponding class labels.}
    \label{fig:resultsImages}
\end{figure*}
\subsection{Classification}
\label{subsection:classification}
Once the vertices have been identified in the feature space for all of the images, we further regularize them to reduce their dispersion. To do this, we linearly interpolate the vertices (0.75) with the average features of all the crops of the corresponding image. These coefficients are fixed through hyperparameter search. The classification of a query image is then done by identifying its closest vertex (Euclidean metric) to one of the vertices of the support images.

\section{Results}
\label{results}

To evaluate our method, we use the standard Mini-ImageNet dataset. We use the classical few-shot setting which considers a new dataset containing 5 classes, with 1-shot and 15 queries per class. We report our results for $10^5$ runs sampled uniformly at random from Mini-ImageNet.  Table \ref{tab:results_miniImageNet} compares the results of our method\footnote{The code to reproduce the results of our experiments is available in the following link: \url{https://github.com/ybendou/few-shot-simplex}.} with those of the state of the art.
\vspace{-1mm}
\begin{table}[!htbp]
    \caption{Results on \textbf{Mini-ImageNet} with 1-shot.}
    \centering
    \begin{tabular}{lc}
    \toprule
         Method & 1-shot 5-ways (\%) \\
         \midrule
         SimpleShot~\cite{wang2019simpleshot}  &$62.85\pm0.20$\\
         Baseline++~\cite{Chen2019} & $53.97\pm0.79$\\
         FEAT~\cite{Ye2018} & $66.78\quad \quad \quad$\\\
       
         Deep EMD~\cite{zhang2020deepemd} & $68.77\pm0.29$\\ 
         PAL~\cite{ma2021partner}& $69.37\pm0.64$ \\
         inv-eq~\cite{rizve2021exploring} & $67.28\pm0.80$\\ 
         CSEI~\cite{li2021learning}& $68.94\pm0.28$\\
         COSOC~\cite{luo2021rectifying} & $69.28\pm0.49$\\
         ASY ~\cite{Bendou2022}& $70.77 \pm 0.06$\\
        \midrule
        \small Ours & \quad $\mathbf{70.90 \pm 0.06\quad}$\\
        \bottomrule
    \end{tabular}
    \label{tab:results_miniImageNet}
\end{table}  


In addition to Figure \ref{fig:QR}, we visualize in Figure \ref{fig:resultsImages} the closest crop of the image to each extracted vertex in the feature space obtained for some images of the dataset. We observe that the vertices are close to objects of interest, one of which often corresponds to the object associated with the image label. The other vertices correspond to objects which are not present in the labels of the dataset. In fact, when a query image is compared to a support image, it is possible to disambiguate the problem by identifying the closest vertices between the two images. 
\begin{figure}[!htbp]
    \centering
    \scalebox{0.9}{
\begin{tikzpicture}

\definecolor{dodgerblue0143213}{RGB}{0,143,213}
\definecolor{lightgray203}{RGB}{203,203,203}
\definecolor{whitesmoke240}{RGB}{240,240,240}

\begin{axis}[
axis line style={whitesmoke240},
tick align=outside,
tick pos=left,
x grid style={lightgray203},
xmajorgrids,
height=5.2cm,
width=8.8cm,
ymin=70.48, 
legend pos=south east,
xtick style={color=black},
y grid style={lightgray203},
ymajorgrids,
ytick style={color=black},
xlabel={\small{$\lambda$}},
ylabel={\small{Performance $(\%)$}},
xmode=log
]
\addlegendentry{\small{Images with one vertex}}

\addplot [line width=16pt, dodgerblue0143213, opacity=0.2, forget plot]
table {%
0.001  70.98
0.01 70.98
0.05 71.1
0.1 71.03
0.15 71.15
0.2 71.11
0.25 71.04
0.3 70.92
0.35 70.88
0.4 70.78
0.45 70.81
0.5 70.81
0.55  70.77
0.6  70.77
0.65  70.77
0.7  70.77
0.75  70.77
0.8  70.77
0.85  70.77
0.9  70.77
0.95  70.77
1  70.77
};

\addplot [ultra thick, dodgerblue0143213]
table {%
0.001  70.98
0.01 70.98
0.05 71.1
0.1 71.03
0.15 71.15
0.2 71.11
0.25 71.04
0.3 70.92
0.35 70.88
0.4 70.78
0.45 70.81
0.5 70.81
0.55  70.77
0.6  70.77
0.65  70.77
0.7  70.77
0.75  70.77
0.8  70.77
0.85  70.77
0.9  70.77
0.95  70.77
1  70.77
};

\addplot [line width=16pt, red, opacity=0.2, forget plot]
table {%
0.001  70.77
1  70.77
};

\addplot [ultra thick, red]
table {%
0.001  70.77
1  70.77
};
\addlegendentry{\small{All images}}

\end{axis}

\end{tikzpicture}}
    \vspace{-4mm}
    \caption{Performance on $10^5$ few-shot classification problems from Mini-ImageNet with one vertex as a function of the regularization parameter.}
   \label{fig:resultsPlot}
\end{figure}
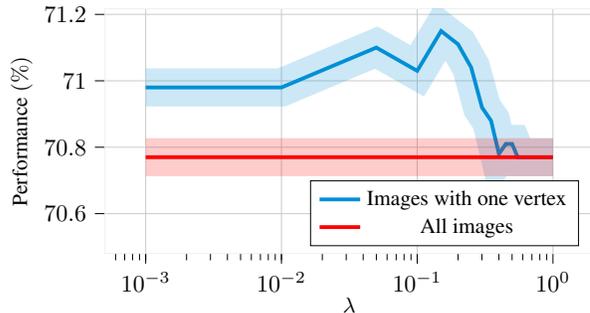
Another way to exploit the vertices extracted by our method is to clean our dataset. Indeed, the images containing more than one object will act as a confusion factor and impact the classification performance. Figure \ref{fig:resultsPlot} presents the performance obtained, when we consider only the examples for which $K=1$ is optimal, as a function of the regularization parameter $\lambda$. When the regularization is high, we come back to the classical case with all the data.

\section{Conclusion}
\label{conclusion}
In this paper, we observed that the disambiguation of the data, by identifying the objects of interest, allows an increase in classification performance in few-shot settings. Furthermore, our method is applicable to any pre-trained feature extractor without retraining it or adapting it to the unseen task. These results are encouraging, and open the door to further improvement when considering problems with multiple class-labeled data, as a similar approach should better identify the common object between multiple examples of the same class.


\bibliographystyle{IEEEbib}
\bibliography{bib}

\end{document}